\documentclass{article}
\usepackage{spconf,amsmath}
\usepackage{booktabs, tabularx, multirow, graphics}
\usepackage{array, amssymb}
\usepackage{enumitem}
\usepackage{graphicx, color}
\usepackage{cite}
\usepackage{xr}
\usepackage{hyperref}
\usepackage[normalem]{ulem}

\DeclareMathOperator*{\argmax}{arg\,max}

\definecolor{dkgreen}{RGB}{0,179,36}
\definecolor{dkred}{RGB}{240,0,0}
\definecolor{dkblue}{RGB}{0,100,240}
\definecolor{dkorange}{RGB}{230,115,0}
\definecolor{pink}{RGB}{255,0,247}

\newcommand{\wv}{W2V2}

\newcommand{\fig}{Fig.}
\newcommand{\figs}{Figs.}
\newcommand{\tab}{Tab.}
\newcommand{\sect}{Sec.}

\title{LAYER-WISE ANALYSIS OF A SELF-SUPERVISED SPEECH REPRESENTATION MODEL}
\name{Ankita Pasad, Ju-Chieh Chou, Karen Livescu}
\address{Toyota Technological Institute at Chicago\\
{\small\texttt{\{ankitap, jcchou, klivescu\}@ttic.edu}}}
\begin{document}
%
\maketitle
\begin{abstract}
\vspace{-.05in}
Recently proposed self-supervised learning approaches have been successful for pre-training speech representation models. The utility of these learned representations has been observed empirically, but not much has been studied about the type or extent of information encoded in the pre-trained representations themselves. Developing such insights can help understand the capabilities and limits of these models and enable the research community to more efficiently develop their usage for downstream applications. In this work, we begin to fill this gap by examining one recent and successful pre-trained model (wav2vec 2.0), via its intermediate representation vectors, using a suite of analysis tools. We use the metrics of canonical correlation, mutual information, and performance on simple downstream tasks with non-parametric probes, in order to (i) query for acoustic and linguistic information content, (ii) characterize the evolution of information across model layers, and (iii) understand how fine-tuning the model for automatic speech recognition (ASR) affects these observations. Our findings motivate modifying the fine-tuning protocol for ASR, which produces improved word error rates in a low-resource setting.\footnote{Codebase: \href{https://github.com/ankitapasad/layerwise-analysis/}{https://github.com/ankitapasad/layerwise-analysis/}}
\end{abstract}
\vspace{-.1in}
\section{Introduction}
\vspace{-.05in}
Self-supervised learning (SSL) techniques 
leverage large-scale unlabeled data to learn meaningful representations~\cite{doersch2015unsupervised,devlin2018bert, radford2018improving}. In such techniques, the unlabeled data is used to design an input and corresponding target output, without any manual annotations. The learned representations are then used as input to a supervised model (and often fine-tuned) for a downstream task. The expected outcome is to either improve downstream task performance or to reduce the amount of labeled data required for training. For the speech domain, various SSL techniques have recently been shown to improve downstream task performance~\cite{oord2018representation, pascual2019learning, schneider2019wav2vec, chung2020generative, wang2020unsupervised, liu2020mockingjay, hsu2021hubert, baevski2020wav2vec, ling2020deep, yang2021superb}. Although new and improved approaches are being proposed at a rapid rate, the pre-trained representations themselves are not well-understood, leaving the development and application of SSL models as a time- and resource-consuming process of trial and error.

\begin{figure}[t]
 \centering
  \includegraphics[width=0.9\linewidth]{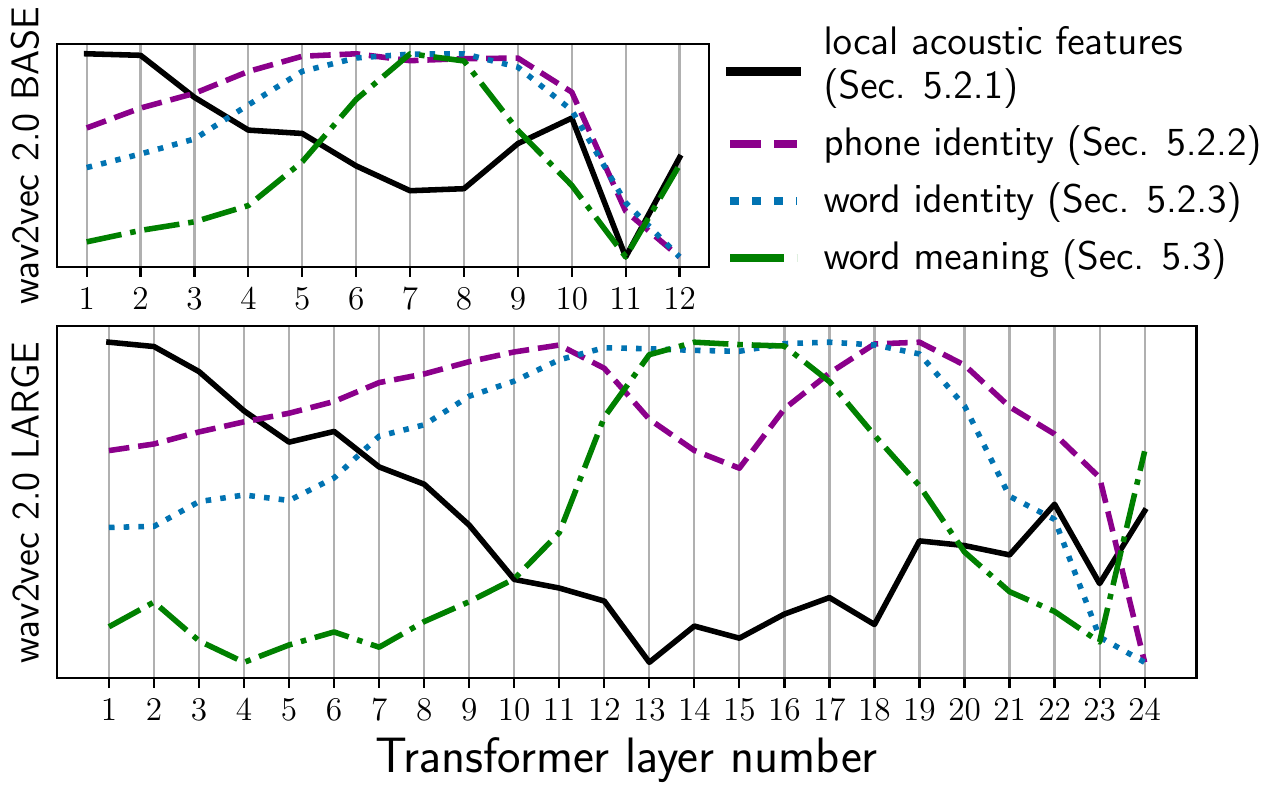}
 
 \vspace{-0.3cm}
 \caption{\it Visualization of properties encoded at different W2V2 layers. The curves measure different metrics on different scales; they are shown together only to compare where major peaks and valleys occur. Details in the indicated sections.}
 \label{fig:all-in-one}
 \vspace{-0.5cm}
\end{figure}

\begin{figure*}
 \centering
 \includegraphics[width=0.9\linewidth]{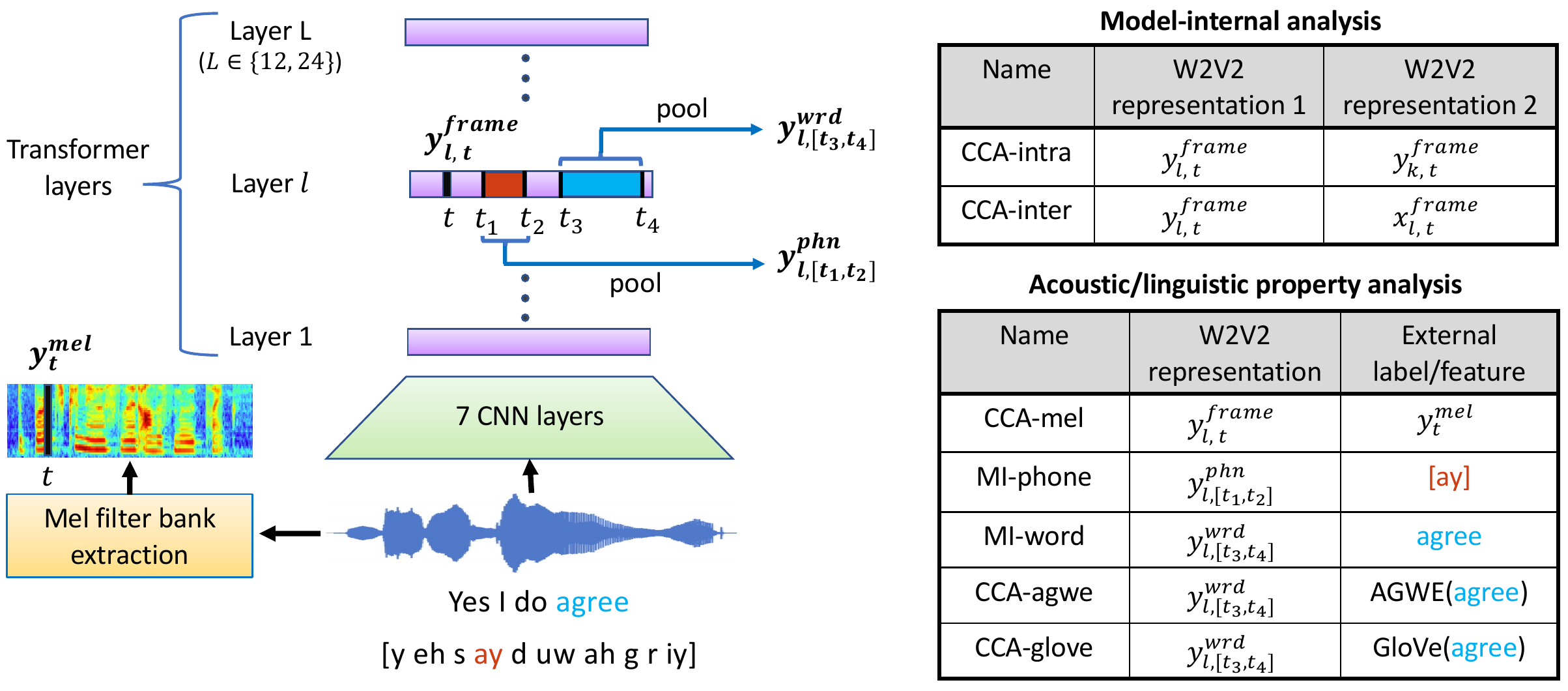}
 
 \vspace{-0.3cm}
 \caption{\it Summary of our analyses using CCA and MI.
 {\it Left:} W2V2 architecture sketch. The Base and Large models have $L = 12$ and $24$ transformer layers respectively.
 {\it Right:} Representations/labels used for each experiment. ``Pool" refers to a pooling operation to combine frame representations into a phone/word segment representation, shown here for a segment corresponding to the phone [ay] and the word ``agree"; in our experiments we use mean pooling. ``AGWE" and ``GloVe" refer to acoustically grounded word embeddings and GloVe word embeddings respectively. See \sect~\ref{sec:setup-details} for details.}
 \label{fig:model}
 \vspace{-0.5cm}
\end{figure*}

We seek to fill this gap by analyzing pre-trained models to understand how the representations evolve across layers, how they relate to a range of linguistic properties, and how they change when fine-tuned for a downstream task. We are especially interested in developing tools to study representations directly, rather than training additional classifiers as probes, to avoid the overhead and unclear dependence on design decisions involved in training classifiers. In this work, we focus our analysis on the open-source wav2vec 2.0 (W2V2) models~\cite{baevski2020wav2vec}, which have been successful for speech recognition~\cite{hsu2021robust,conneau2020unsupervised,baevski2021unsupervised} and translation~\cite{wang2021large}. 
Our main findings are:

\begin{itemize}[leftmargin=*,noitemsep,nolistsep]
    \item The W2V2 transformer layers follow an autoencoder-style behavior, where as we go deeper into the model, the representation starts deviating from the input speech features followed by a reverse trend where even deeper layers become more similar to the input, as if reconstructing the input.
    \item The layer-wise evolution of the representations follows an acoustic-linguistic hierarchy, where the shallowest layers encode acoustic features, followed by phonetic, word identity, and word meaning information (and then followed by a reverse trend as described above), as illustrated in \fig~\ref{fig:all-in-one}.
    \item Fine-tuning the model for ASR breaks the autoencoder-style behavior in the final few layers, which accordingly also get better at encoding word identity.
    \item The final convolutional (CNN) layers and initial transformer layers are highly correlated with mel spectrogram features, suggesting that the model learns to extract features similar to human-engineered ones.
    \item The model encodes some word meaning information.
    \item The last two layers often defy the previous layers' trends.
    \item  A fine-tuning protocol, designed based on these findings, improves ASR performance in low-resource settings.
\end{itemize}
 
\vspace{-.1in}
\section{Related work}
\vspace{-.1in}
There has been extensive work on analyzing supervised speech models~\cite{belinkov2019analysis,palaskar2019learned,prasad2020accents}, but research on analyzing SSL models has been limited. Some very recent work has explored the phonetic, paralinguistic, and semantic content in SSL models using  classifier probes~\cite{hsu2021hubert,baevski2021unsupervised,ma2021probing,shah2021all} and relationships between models with different training objectives and architectures~\cite{chung2020similarity}.
The 2021 Zero Resource Speech Benchmark~\cite{nguyen2020zero} introduces zero-shot analysis datasets and metrics to evaluate the ability of SSL speech representations to encode different levels of linguistic information. While we share much of the motivation of \cite{ma2021probing,shah2021all,nguyen2020zero}, we focus on layer-wise analysis of a range of acoustic-linguistic content using lightweight methods that don't rely on training classifiers or collecting any additional labels for analysis, making it easier to scale.

Layer-wise analysis of linguistic structure has also been done before for visually grounded speech~\cite{chrupala2017representations} and SSL text models~\cite{tenney2019bert}.
Our methods of canonical correlation analysis (CCA) and discrete mutual information (MI) estimates are closest to Voita {\it et al.}'s work on text models~\cite{voita2019bottom}. MI has also been used to analyze supervised ASR models~\cite{prasad2020accents}. Unlike prior work, we apply these methods to the analysis of the relationship between representations and both discrete labels and continuous embeddings, and between representations from pre-trained and fine-tuned models. To our knowledge, this is the first work to analyze an SSL speech model on a range of linguistic properties using non-parametric probes.

\section{Analysis Methods}
\label{sec:methods}
\vspace{-.075in}
\fig~\ref{fig:model} sketches the W2V2 model structure and the representations used in many of our analyses.

{\bf Canonical Correlation Analysis.} CCA~\cite{harold1936relations} is a statistical technique that measures the relationship between two continuous-valued random vectors as represented by the maximum correlations between their linear projections. CCA has been previously used as a measure of similarity to compare representations within and across neural network models~\cite{raghu2017svcca, kornblith2019similarity, voita2019bottom}. Here we use it in the same way, and also to measure the similarity between a layer representation and another vector, such as word embeddings or acoustic features.

CCA takes $n$ pairs of vectors $\{(x_1, y_1), ..., (x_n, y_n)\}$, sampled from the random vectors (or ``views") $X\in\mathbb{R}^{d_1}, Y\in\mathbb{R}^{d_2}$, as input and returns a correlation score as a measure of similarity between the two views.
The solution can be defined iteratively as follows: First we define the directions of maximum correlation between linear projections of $X$ and $Y$: $v_1, w_1 = \argmax_{v, w} \text{corr}(v^TX, w^TY)$. The subsequent directions $v_i, w_i \; \forall i \in [2, k]$, $k = \min(d_1, d_2)$, maximize the same correlation subject to each new projection being uncorrelated with others in the same view.

In standard CCA the {\emph canonical correlation} $\textrm{CCA}(X, Y)$ is the sum (or mean) of the correlations $\rho_i = \text{corr}(v_i^TX, w_i^TY)$. We use a variant, {\it projection-weighted CCA (PWCCA)}~\cite{morcos2018insights}, which computes a weighted mean of the $\rho_i$s, with higher weights for directions accounting for a higher proportion of the input. PWCCA has been found to be more robust to spurious correlations in the data. Since PWCCA is asymmetric, we report the mean of the two quantities $\textrm{CCA}(X, Y)$ and $\textrm{CCA}(Y, X)$. Henceforth we refer to this average as the ``CCA similarity", and it has a maximum value of 1.

As illustrated in \fig~\ref{fig:model}, we use PWCCA to measure similarity between the W2V2 layer representations and various continuous-valued quantities of interest, either (i) from a different layer of the same model ({\it CCA-intra}), (ii) from a fine-tuned version of the model ({\it CCA-inter}), or (iii) from an external representation. For the third type of analysis we use mel filter bank features ({\it CCA-mel}), acoustically grounded word embeddings~\cite{settle2019acoustically} ({\it cca-agwe})\footnote{AGWEs are trained to be close to acoustic embeddings of the corresponding words, so we expect they encode mainly acoustic-phonetics.} and GloVe word embeddings~\cite{pennington2014glove} ({\it cca-glove}) as ways to assess the local acoustic, word-level acoustic-phonetic, and word meaning information encoded in the W2V2 representations respectively.

{\bf Mutual information.}
While CCA is natural for relating continuous-valued vectors, we use mutual information (MI) to measure 
dependence between the representations, $y^{phn}$ or $y^{wrd}$ from \fig~\ref{fig:model}, and the corresponding phone (\emph{MI-phone}) or word (\emph{MI-word}) label. We cluster the continuous-valued representations to obtain discrete clusters, as in~\cite{voita2019bottom, prasad2020accents}.

{\bf Word discrimination.}
({\it word-disc}) is the task of detecting whether two speech segments correspond to the same or different words~\cite{carlin2011rapid} and is commonly used to evaluate acoustic word embeddings and other acoustic representations~\cite{kamper2015unsupervised,hu2020multilingual,algayres2020evaluating,jacobs2021acoustic}. We follow a typical evaluation protocol, where we label a pair of segments as ``same word" if the cosine similarity between their word-level representations is above some threshold, and measure performance via the average precision as the threshold is varied. We use this task-specific measure primarily to corroborate our findings from MI-word.

{\bf Word similarity tasks.} 
We perform a suite of 11 standard word similarity tasks ({\it word-sim})~\cite{faruqui2014community} as an additional measure of word meaning information.\footnote{https://github.com/vecto-ai/word-benchmarks} We extract context-independent word embeddings from W2V2, as described in \sect~\ref{sec:setup-details}. The semantic similarity score for each word pair is measured as the cosine similarity between these embeddings. Performance is measured as the Spearman's $\rho$ correlation between these scores and the human similarity judgements.

\vspace{-.1in}
\section{Experimental Setup}
\vspace{-.05in}
\subsection{Representation Learning Model}
\vspace{-.05in}
\label{sec:rep-model}
The W2V2 model~\cite{baevski2020wav2vec} maps raw waveforms to higher-level contextual features via a set of convolutional layers followed by self-attention (transformer) layers, as shown in \fig~\ref{fig:model}, and is trained with a contrastive objective that measures the ability of the model to differentiate between a true masked input segment and a set of distractors. The self-attention layers in the transformer allow the model to encode information from the context surrounding a given masked segment.

We analyze three W2V2 variants: (i) {\it Base}: 12 layers, trained on 960 hours of LibriSpeech~\cite{panayotov2015librispeech},
(ii) {\it Large-960}: 24 layers, trained on 960 hours of LibriSpeech,
(iii) {\it Large-60k}: 24 layers, trained on 60k hours of LibriVox~\cite{kahn2020libri}.
We also analyze W2V2 fine-tuned for ASR with 10 minutes ({\it ft-10m}), 100 hours ({\it ft-100h}), and 960 hours ({\it ft-960h}) of labeled LibriSpeech.\footnote{All models are downloaded from the wav2vec 2.0 repository: https://github.com/pytorch/fairseq/blob/master/examples/wav2vec}
Fine-tuning consists of adding a randomly initialized linear layer to the pre-trained model, and then training with character-level connectionist temporal classification (CTC) loss~\cite{graves2006connectionist}, while keeping the CNN layers frozen~\cite{baevski2020wav2vec}. We refer to this as the ``standard approach" in Sec.~\ref{sec:asr}. 

\vspace{-.15in}
\subsection{Setup Details}
\label{sec:setup-details}
\vspace{-.05in}

We perform all experiments on LibriSpeech. The sampled utterances (details in \tab~\ref{tab:data-specs}) are passed through each W2V2 model, and the outputs from all layers are extracted.
Random masking is turned off except for experiments analyzing the effect of masking (\sect~\ref{sec:masking-effect}).

\begin{table}[]
\small
\begin{center}
\begin{tabular}{l|ll}
{\bf Experiment}                               & {\bf \# labels}          & {\bf \# representation examples}
\\ \hline 
\begin{tabular}[c]{@{}l@{}}CCA-intra, \\ CCA-inter, \\ CCA-mel\end{tabular} & n/a            & 150k frames            \\ \hline
\begin{tabular}[c]{@{}l@{}} CCA-agwe, \\ CCA-glove \end{tabular}                           & 2.7k words         & 4.8k word segments           \\ 
\hline
\multirow{2}{*}{MI-phone}                         & \multirow{2}{*}{39 phones} & train: 187k phone segments \\ 
                                     &              & dev: 7.6k phone segments          \\ 
                                     \hline
\multirow{2}{*}{MI-word}                         & \multirow{2}{*}{500 words} & train: 427k word segments  \\ 
                                     &              & dev: 6.9k word segments         \\ 
                                     \hline
\multirow{2}{*}{word-disc}                                & \multirow{2}{*}{300 words}       & 2.4k word segments \\ & & (2.9M pairs)   \\ 
\end{tabular}
\vspace{-.05in}
\caption{\it Data subsets curated for our analysis. We repeat each experiment on four sample sets. The numbers here are averages across the four sets. For MI experiments, the train subsets are used to define the clustering. For word-disc, we use words that are at least 5 characters and 500ms long.}
\vspace{-0.85cm}
\label{tab:data-specs}
\end{center}
\end{table}

\textbf{Representation extraction}: 
We use LibriSpeech alignments generated using the Montreal forced aligner~\cite{mcauliffe2017montreal, lugosch2019speech} to define phone and word segments. As illustrated in \fig~\ref{fig:model}, word-level representations $y^{wrd}$ are obtained by averaging the frame representations of all frames in a given word segment. Phone-level representations $y^{phn}$ are obtained by averaging the frame representations of the central third of each phone segment; the first and last third are discarded to reduce co-articulation effects. These segment representations are used for all experiments in \tab~\ref{tab:data-specs} except the first row. The context-independent embedding (used for the {\it word-sim} experiments) for each word is computed by averaging the $y^{wrd}$ representations across all the instances of that word in train-clean.\footnote{We also tried weighted mean pooling, using averaged attention weights from all the attention heads, which produced similar results.}

\textbf{Mel filter bank features}: 
80-dimensional mel filter bank (fbank) features are extracted using a frame length of 25ms and an overlap of 10ms. In order to make the W2V2 representations comparable to the fbank features, we compute moving averages of CNN features or downsample fbank features as needed to ensure their strides and receptive fields match.

\textbf{Discrete cluster IDs}: For MI experiments, we discretize the continuous-valued W2V2 representations. Specifically, we cluster a set of phone/word-level representations sampled from the train-clean LibriSpeech split with roughly the same number of examples of each label,\footnote{Similar trends are obtained when the chosen instances are uniformly sampled from the data instead.} using mini-batch k-means with k=500 for MI-phone and k=5000 for MI-word, and assign each development set example to the nearest cluster.

\vspace{-.1in}
\section{Findings}
\label{sec:findings}
\vspace{-.1in}
We present results for experiments done on the dev-clean split on some of the W2V2 variants (base, base-ft-960, large-60k, large-60k-ft-960); the findings generalize to dev-other and to the large-960 model unless stated otherwise. We 
analyze pre-trained models in \sect~\ref{sec:cca-intra}-\ref{sec:word-sim} and their fine-tuned counterparts in \sect~\ref{sec:ftune}.\footnote{The \sect~\ref{sec:cca-intra}-\ref{sec:word-sim} figures combine both pre-trained and fine-tuned models.} Each plot below gives the mean of the relevant measure across the four sample sets; typical variation across the sets is $< 0.02$ for CCA measures, $< 0.07$ for MI, and $< 2\%$ for word-disc. We refer to the output of transformer layer $l$ as the representation at layer $l$ and the output of the CNN feature encoder as layer $0$ or "local features".
\vspace{-.15in}
\subsection{How do the representations evolve across layers?}
\label{sec:cca-intra}
\begin{figure}[h]
\begin{minipage}[b]{1.0\linewidth}
\small

\vspace{-0.5cm}
 \centering
 \centerline{\includegraphics[width=8cm, trim=0 185 0 0, clip]{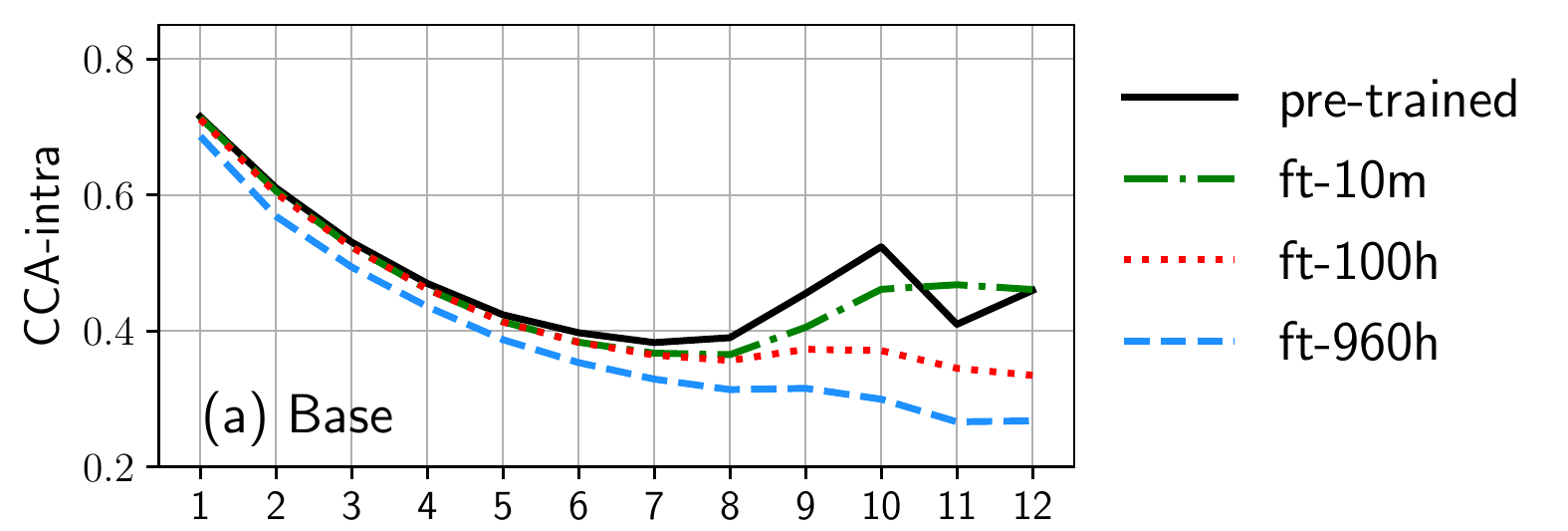}}
\end{minipage}
\begin{minipage}[b]{1.0\linewidth}

\vspace{-0.05cm}
\footnotesize
 \centering
 \centerline{\includegraphics[width=8cm, trim=0 168 0 0, clip]{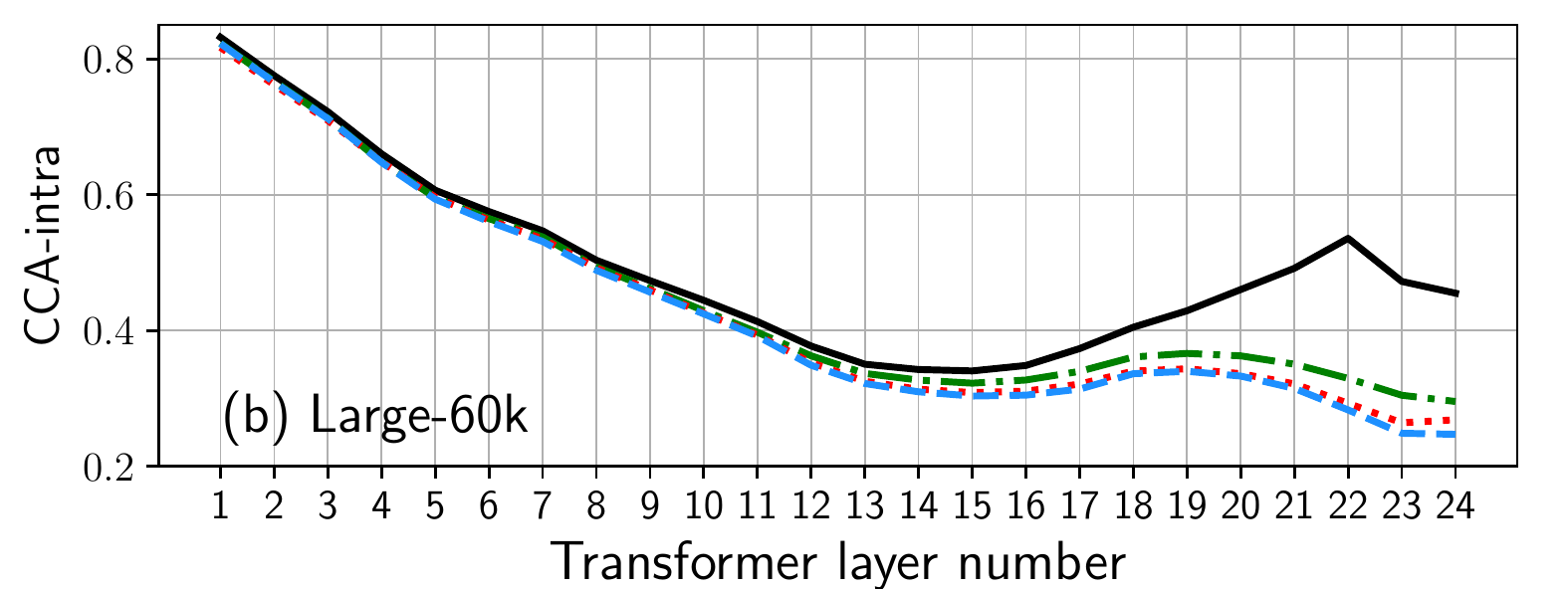}}
\end{minipage}

\vspace{-0.3cm}
\caption{\it CCA similarity with local features.}
  \label{fig:cca-intra}
\vspace{-0.3cm}
\end{figure}

\noindent In \fig~\ref{fig:cca-intra} we compare (via CCA similarity) the transformer layer representations with the ``local features" extracted by the CNN module (layer $0$). We see that the pre-trained model (solid black curve) follows an autoencoder-style behavior, where as we go deeper into the model, the representation starts deviating from the input features, followed by a reverse trend where even deeper layers become more similar to the input, as if reconstructing the input (although this trend seems to break for the last two layers; see \sect~\ref{sec:masking-effect}). Since the training objective is to distinguish the masked input segment from distractors, it is natural for the final layers to have similar properties to the input. A similar behavior, referred to as context-encoding and reconstruction, has been observed for the BERT text model~\cite{voita2019bottom}, where the objective is based on masked reconstruction rather than contrastive prediction.

\subsection{Where is acoustic/linguistic information encoded?}
\label{sec:linguistic-property}
\vspace{-.05in}
Next we consider how certain properties are encoded in different layers. As a reminder, all our experiments are performed on features extracted locally from a short span of frames (frame/phone/word-level). Any increase in ``information" across layers for these local representations is possible due to contextualization from the self-attention layers that enable each frame-level output to access the whole utterance. For the same reason, any decrease in ``information" across layers could be attributed to de-localization, i.e.~the information is no longer localized to the frame/phone/word segment.
\vspace{-.13in}
\subsubsection{Frame-level acoustic content}
\vspace{-.05in}
\begin{figure}[h]
\begin{minipage}[b]{1.0\linewidth}
\small

\vspace{-0.4cm}
 \centering
 \footnotesize
 \centerline{\includegraphics[width=9cm, trim=0 221 0 2, clip]{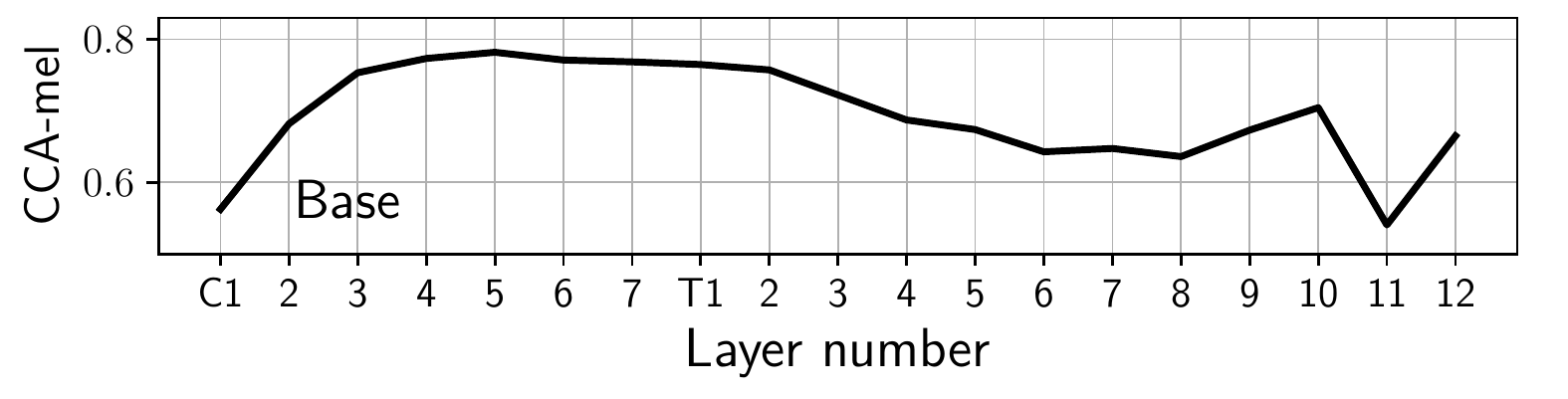}}
\end{minipage}

\vspace{-0.3cm}
\caption{\it CCA similarity between layer representations and fbank; C$i$: CNN layer $i$, T$j$: transformer layer $j$.}
\label{fig:cca-mel}
\vspace{-0.4cm}
\end{figure}

\noindent \fig~\ref{fig:cca-mel} shows the layer-wise CCA similarity between fbank vectors and Base model layers. For the first few layers the correlation increases with depth. The Large models follow a similar curve, with high correlation for layers C4-T2 ($> 0.75$). We can infer that the model learns to compute features much like fbank, suggesting a potential simplification to W2V2 to take fbank as input (which we leave to future work). 

\vspace{-.13in}
\subsubsection{Phonetic information}
\vspace{-.05in}
\begin{figure}[h]
\begin{minipage}[b]{1.0\linewidth}
\small

\vspace{-0.4cm}
 \centering
 \centerline{\includegraphics[width=8.5cm, trim=0 240 0 2, clip]{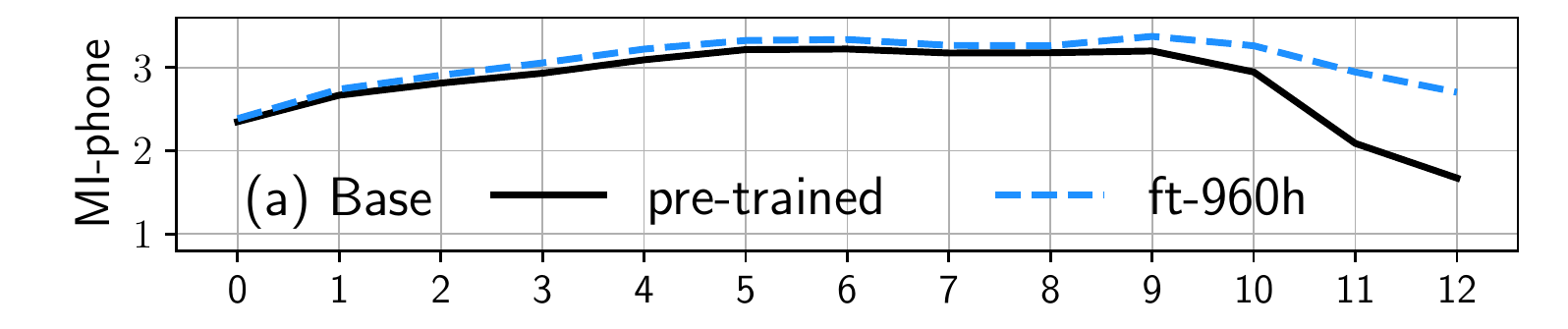}}
\end{minipage}
\begin{minipage}[b]{1.0\linewidth}

\vspace{-0.17cm}
\small
 \centering
 \centerline{\includegraphics[width=8.5cm, trim=0 125 0 116, clip]{images-large/small.pdf}}
\end{minipage}
\begin{minipage}[b]{1.0\linewidth}

\vspace{-0.1cm}
\small
 \centering
 \centerline{\includegraphics[width=8.5cm, trim=0 240 0 2, clip]{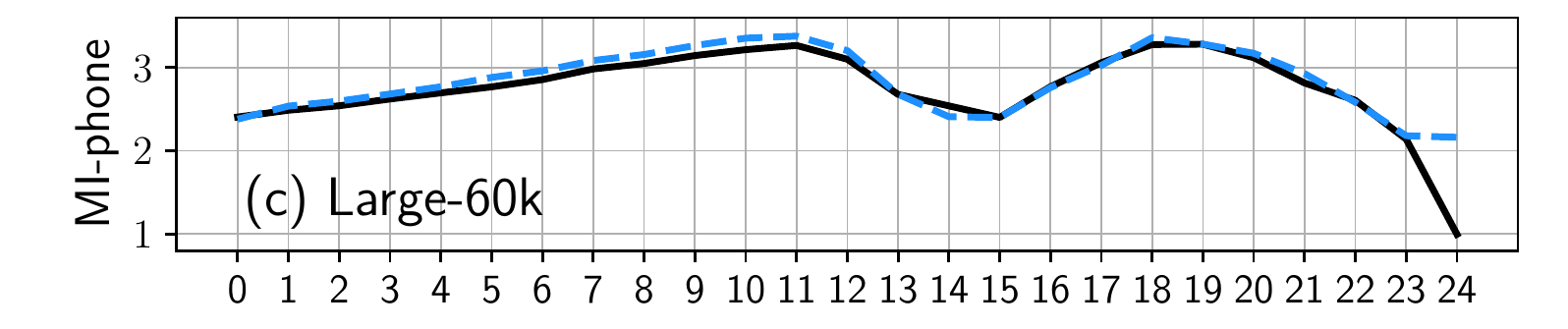}}
\end{minipage}
\begin{minipage}[b]{1.0\linewidth}

\vspace{-0.15cm}
\footnotesize	
 \centering
 \centerline{\includegraphics[width=8.5cm, trim=0 105 0 116, clip]{images-large/vox.pdf}}
\end{minipage}

\vspace{-0.3cm}
\caption{\it MI with phone labels (max:~3.6) and CCA similarity with AGWE.}
\label{fig:mi-phone}
\vspace{-0.3cm}
\end{figure}

\noindent We measure the phonetic information encoded in the pre-trained model in two ways, MI-phone and CCA-agwe\footnote{We use AGWEs trained on LibriSpeech similarly to~\cite{settle2019acoustically}.} (see \sect~\ref{sec:methods}), both shown in \fig~\ref{fig:mi-phone}. We expect AGWEs to encode mostly phonetic information, and indeed the phone and AGWE curves in \fig~\ref{fig:mi-phone} follow broadly similar trends.

We notice that phonetic information appears to be most salient around layer 6-7 for Base (similarly to \cite{hsu2021hubert}, which includes a similar experiment). For Large-60k (\fig~\ref{fig:mi-phone}), however, layers 11 and 18/19 appear equally adept at encoding phonetic information, with a drop in between.

\vspace{-.125in}
\subsubsection{Word identity}
\vspace{-.05in}
\begin{figure}[h]
  \begin{minipage}[b]{1.0\linewidth}
\small

\vspace{-0.5cm}
 \footnotesize
 \centerline{\includegraphics[width=8.5cm, trim=0 240 0 2, clip]{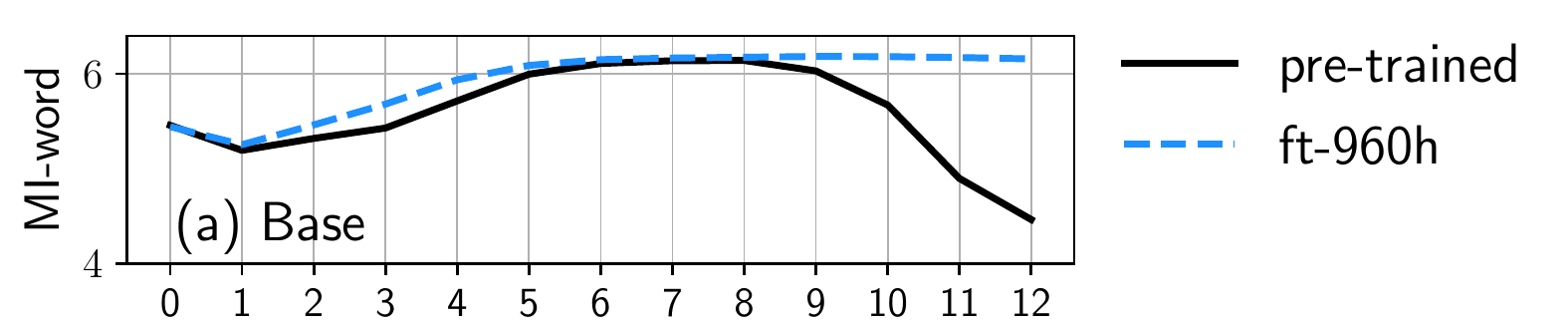}}
\end{minipage}
 \begin{minipage}[b]{1.0\linewidth}
\small

 \footnotesize
 \centerline{\includegraphics[width=8.5cm, trim=0 225 0 2, clip]{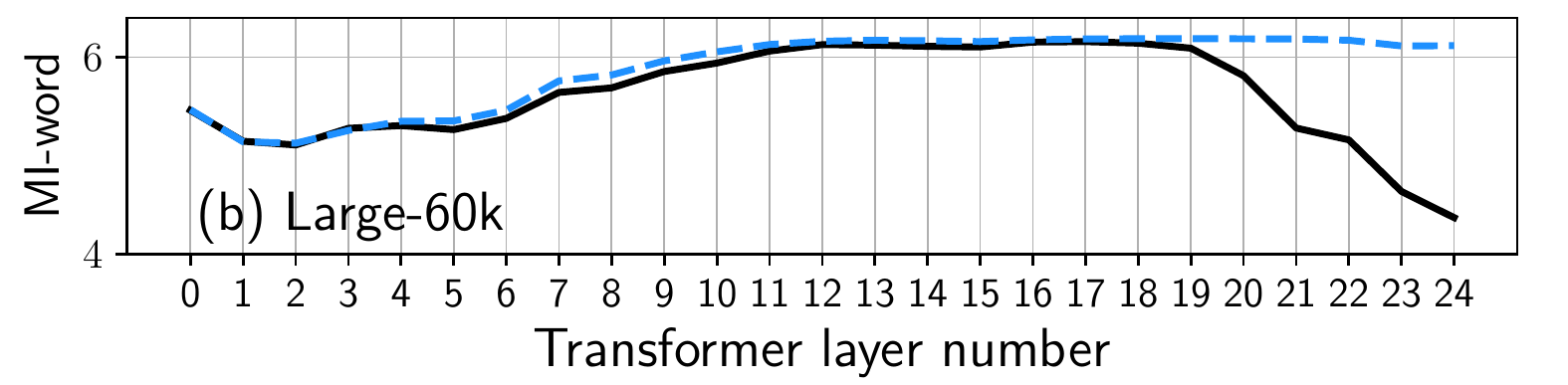}}
\end{minipage}

\vspace{-0.4cm}
  \caption{\it MI with word labels (max:~6.2).}
  \label{fig:mi-word}
 \vspace{-0.2cm}
\end{figure}

\noindent \fig~\ref{fig:mi-word} shows the mutual information between the layer representations and word labels. For Base, the trends are similar to those of MI-phone (\fig~\ref{fig:mi-phone}a). For Large-60k (\fig~\ref{fig:mi-word}b), word identity appears to be encoded similarly well by layers 12 to 18, without the dip seen in the MI-phone curve. 

\begin{figure}[h]
   \vspace{-0.25cm}
  \begin{minipage}[b]{1.0\linewidth}
\small

\vspace{-0.0cm}
 \centering
 \footnotesize
 \centerline{\includegraphics[width=8.5cm, trim=0 221 0 2, clip]{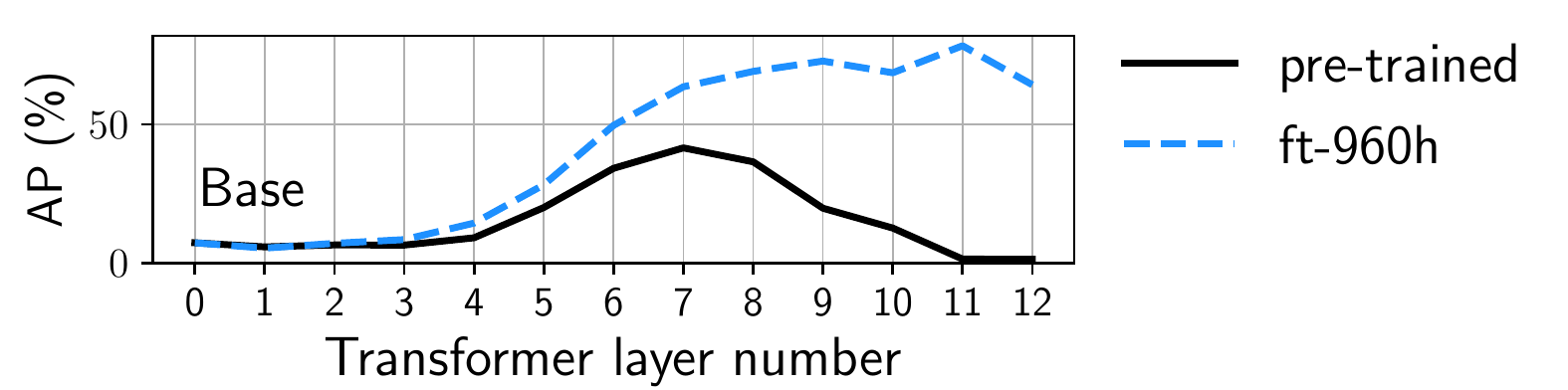}}
\end{minipage}

\vspace{-0.35cm}
  \caption{\it Average precision (AP) for word discrimination.}
  \label{fig:word-disc}
  \vspace{-0.3cm}
\end{figure}

As another measure of word identity content, \fig~\ref{fig:word-disc} shows word discrimination performance, which follows a similar trend to MI-word (\fig~\ref{fig:mi-word}a). We experiment with all of the W2V2 variants (including ones not shown here), and find that MI-word and word-disc are always highly correlated.

\vspace{-.1in}
\subsubsection{Evidence for the most contextual layers}
\vspace{-.05in}
For the Large-60k model, the curves measuring acoustic-phonetic information (\figs~\ref{fig:cca-intra}b,~\ref{fig:mi-phone}c,~\ref{fig:mi-phone}d) all have a dip around layers 13-17 (see also \fig~\ref{fig:all-in-one}). These are also the same layers that seem to have the {\it most} word content (\fig~\ref{fig:mi-word}b). This suggests that around these layers, the model may be extracting the most contextual and high-level information, and retaining less lower-level information like phonetic content. Beyond these layers, the model enters the reconstruction phase, thus encoding more local representations at even deeper layers.

The Base model does not have the same significant inter-mediate drop for phonetic content (\figs~\ref{fig:all-in-one},~\ref{fig:mi-phone}a,~\ref{fig:mi-phone}b) as does Large-60k, which could indicate less contextualization. In experiments on Large-960 (not shown here), the MI-phone and CCA-agwe scores do not show this drop either, implying that this effect is the result of the larger training set of Large-60k, and not its larger model size.
\vspace{-.0in}
\subsection{Does the pre-trained model learn word meaning?}
\label{sec:word-sim}
\begin{figure}[h]
    \begin{minipage}[b]{1.0\linewidth}
\small

\vspace{-0.5cm}
 \centering
 \footnotesize
 \centerline{\includegraphics[width=8.5cm, trim=0 221 0 2, clip]{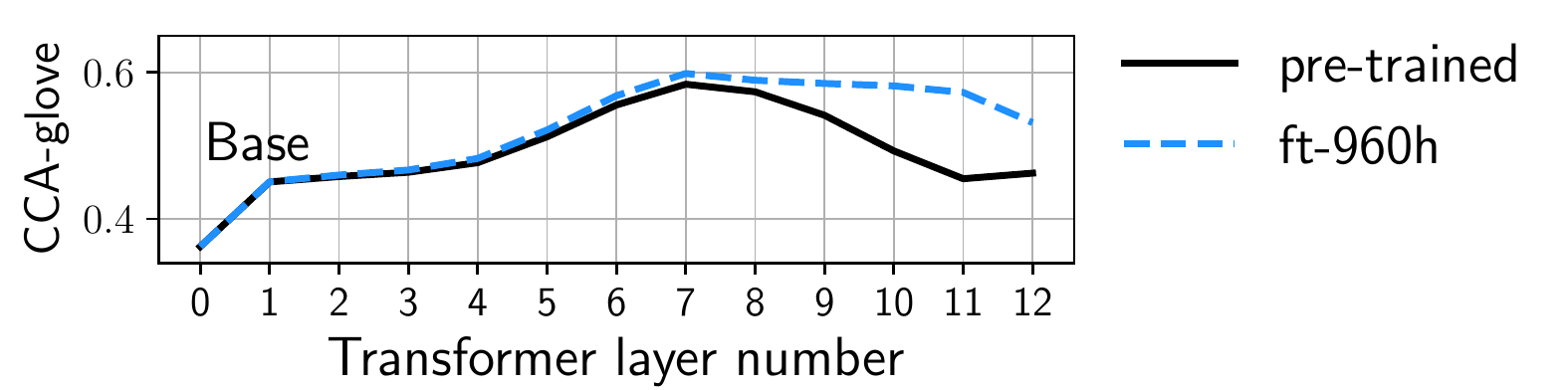}}
 \end{minipage}

\vspace{-0.2cm}
  \caption{{\it CCA similarity with GloVe embeddings.}}
  \label{fig:cca-glove}
  \vspace{-0.3cm}
\end{figure}
\noindent While some linguistic properties seem essential for the model to learn to solve the self-supervised task, it is not obvious that word meaning is one such property. We probe for word meaning in W2V2 by measuring the CCA similarity between word segment representations and the popular text-based GloVe embeddings~\cite{pennington2014glove}, shown in \fig~\ref{fig:cca-glove}. These plots (also a part of \fig~\ref{fig:all-in-one}) provide further evidence that the central layers (7-8 for Base and 14-16 for Large-60k) encode the most contextual information. Note that these curves have a narrower plateau of maximum performance around these layers than the MI-word curves (\fig~\ref{fig:mi-word}), suggesting that the most contextual layers are better at encoding word meaning while the peripheral layers are good at encoding lower-level linguistic content but not meaning.

\begin{figure}[h]
  \vspace{-0.2cm}
  \centering
  \includegraphics[width=\linewidth]{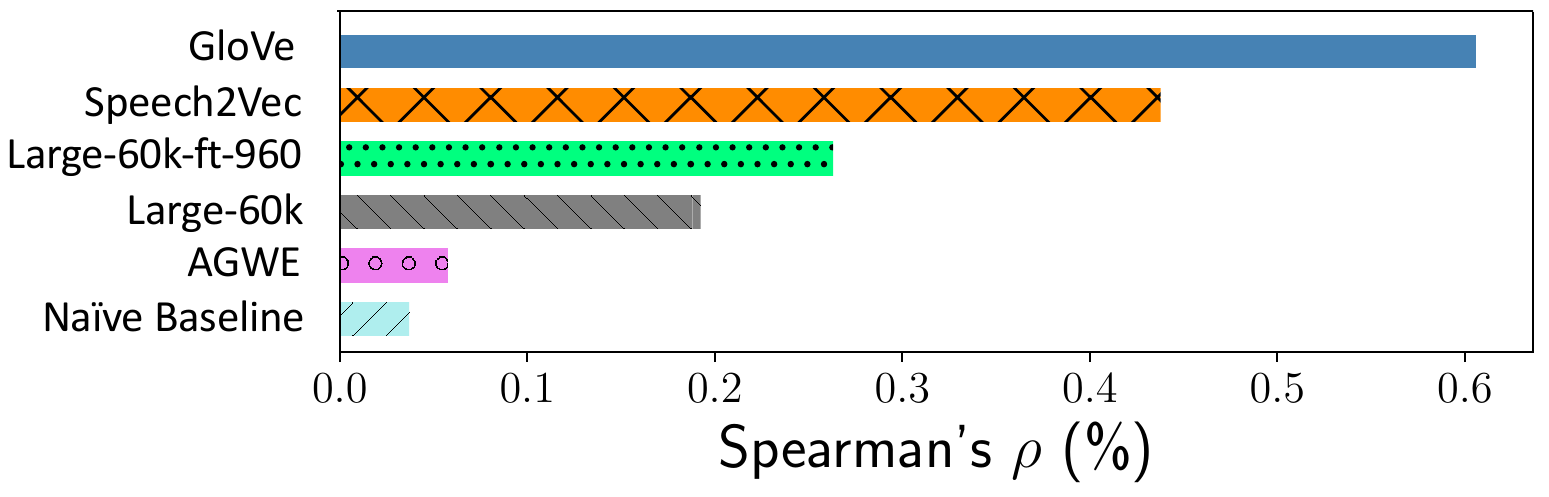}
  
  \vspace{-0.4cm}
  \caption{\it Word similarity performance (mean across tasks).}
  \label{fig:word-sim}
  \vspace{-0.2cm}
\end{figure}
To further calibrate our measure of semantic information, we evaluate the W2V2 representations on standard word similarity benchmarks, as described in \sect~\ref{sec:methods}. \fig~\ref{fig:word-sim} reports the performance of the best layers for Large-60k and Large-60k-ft-960. The best performance for both models occurs at layer 15, which again agrees with our hypothesis that layers 14-16 contain the most semantic information.

We also present two baselines:
(i) the {\it naive baseline} defines word distance as the character edit distance for each word pair; this baseline has non-trivial performance when orthography is a helpful clue
(ii) the {\it AGWE baseline} uses AGWEs in place of the W2V2 representations, and may succeed for word pairs where acoustic-phonetic similarity correlates with semantic similarity.
We also include two models that are trained specifically to encode semantics:
(i) Speech2Vec~\cite{chung2018speech} learns word embeddings from speech using an approach similar to word2vec~\cite{mikolov2013distributed} and is trained on LibriSpeech, and
(ii) GloVe embeddings~\cite{pennington2014glove}.
Since W2V2 is not trained with an explicit semantic criterion, it is not surprising that it is outperformed by Speech2Vec and GloVe. It is interesting, however, that W2V2 representations perform better than 
the non-semantic baselines, suggesting that some meaning is being encoded.
\vspace{-.075in}
\subsection{How does fine-tuning affect the above observations?}
\label{sec:ftune}
\begin{figure}[h]
  \begin{minipage}[b]{1.0\linewidth}
\small

\vspace{-0.5cm}
 \centering
 \centerline{\includegraphics[width=8cm, trim=0 160 0 0, clip]{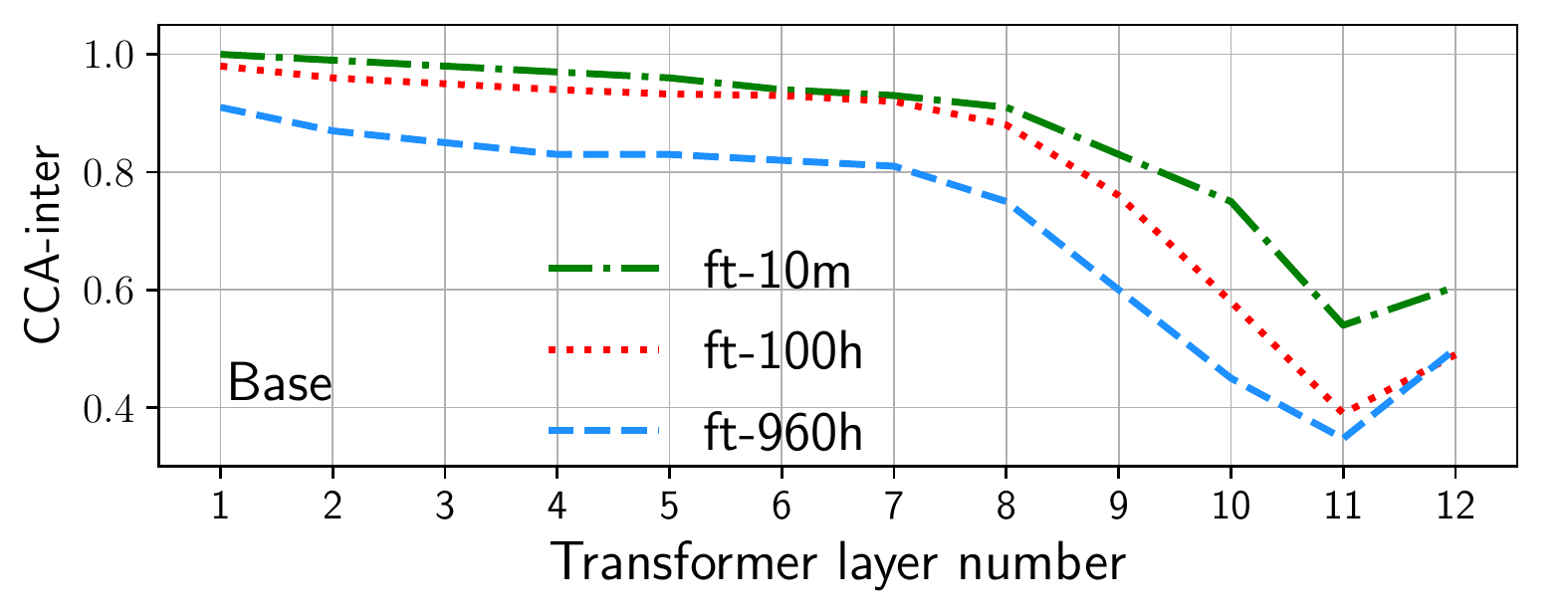}}
\end{minipage}

\vspace{-0.5cm}
  \caption{\it CCA similarity between each layer of a pre-trained model and the same layer of fine-tuned models.}
  \label{fig:cca-inter}
  \vspace{-0.25cm}
\end{figure}

\noindent We see in CCA-intra, \fig~\ref{fig:cca-intra}, that fine-tuning breaks the autoencoder-style behavior. After fine-tuning for ASR, the deeper layers that were originally reconstructing the input are now diverging from the input, and presumably learning more task-specific information. We also see from \fig~\ref{fig:cca-inter} that the higher layers 
change the most in fine-tuning, suggesting that the pre-trained model may not serve as a good initialization of these top layers for ASR. This finding suggests re-initializing these layers before fine-tuning, as has been recently discovered for BERT~\cite{zhang2020revisiting}. We design a fine-tuning experiment to validate this idea, described in \sect~\ref{sec:asr}. 

MI-word consistently improves across the top layers (19-24) after fine-tuning (\fig~\ref{fig:mi-word}). The same does not always hold for phone identity (\fig~\ref{fig:mi-phone}). These results indicate that, as might be expected, fine-tuning with character-level CTC loss is more directly related to the word identity than to phone identity. For the semantic measures (CCA-glove and word-sim) as well we don't see the same large improvements as for MI-word, again as may be expected since ASR does not necessarily require high-quality word meaning representations.
\vspace{-.1in}
\subsection{What about those peculiar last two layers?}
\label{sec:masking-effect}
\vspace{-.05in}
We see a peculiar pattern in most of the CCA similarity curves for pre-trained W2V2 models, where at least one of the last two layers fails to follow the trend of the previous layers. We find that this peculiarity disappears when we turn random masking on and consider only the representations of the masked segments. Moreover, the phonetic and word content, as measured by MI, improves for the last two layers (while reducing for the rest) when working with the representations of masked segments. This finding suggests that the representations of the final two layers are more meaningful when the input segment is masked. Furthermore, this discrepancy is not present in the fine-tuned models, suggesting that this effect is connected to the training objective, but the exact relationship is unclear. We also note that this peculiarity has been observed for local representations extracted from BERT~\cite{ethayarajh2019contextual}. 

\vspace{-.11in}
\section{Practical Implications for ASR}
\label{sec:asr}
\vspace{-.1in}

We have noted that the last few layers of W2V2 change the most during fine-tuning (\fig~\ref{fig:cca-inter}), and that the linguistic content that should be helpful for ASR is less well represented in the final few layers (\figs~\ref{fig:mi-phone}a,~\ref{fig:mi-word}a). Based on these observations, we hypothesize that some of these final layers do not provide a good initialization for the task. To test this hypothesis we modify the ``standard approach" by re-initializing the top layer(s) before fine-tuning. We conduct all ASR experiments using the SpeechBrain toolkit~\cite{ravanelli2021speechbrain}. We experiment with \wv-base and find that re-initializing the final 1-3 layers indeed outperforms the standard approach of initializing all layers from the pre-trained model (\tab~\ref{tab:wer}), with large improvements when fine-tuning on the 10-minute training set and minor improvements for larger training sets.

\begin{table}[]
\centering
\begin{tabular}{cc|cc}
\multirow{2}{*}{\textbf{train set}} & \multirow{2}{*}{\textbf{$n$}} & \multicolumn{2}{c}{\textbf{standard \textrightarrow{} re-init }12-$n$ \textbf{layers}} \\ \cline{3-4}
                                    &                             & \textbf{test-clean}      & \textbf{test-other}     \\ \hline
10m                                 & 9                           & 49.0 \textrightarrow{} 44.1               & 56.7 \textrightarrow{} 51.8              \\
1h                                  & 11                          & 20.3 \textrightarrow{} 19.8               & 29.8 \textrightarrow{} 29.3              \\
10h                                 & 11                          & 11.3 \textrightarrow{} 10.9               & 20.6 \textrightarrow{} 19.4             
\end{tabular}
\caption{\it Word error rates (\%) for the modified fine-tuning protocol for the Base model, using the best value of $n$ based on dev-clean performance, compared to standard fine-tuning. \hspace{-.05in}{\it A\textrightarrow{}B} indicates that standard fine-tuning produces WER {\it A}, and the proposed protocol produces WER {\it B}.} 
\label{tab:wer}
\vspace{-.2in}
\end{table}

\vspace{-.1in}
\section{Conclusion}
\vspace{-.1in}
We have presented a set of analyses to assess the layer-specific information in pre-trained speech representations, applied to wav2vec 2.0 models. 
We find that various acoustic and linguistic properties tend to be encoded in different layers, and the pre-trained model follows an autoencoder-style behavior. We also find that the model encodes some non-trivial word meaning information, although more work is needed to determine the nature of the semantic content. We corroborate most of our findings with multiple analytical measures and certain downstream tasks. Such analyses can help understand the abilities and limitations of models trained without external supervision, and also help direct research toward additional useful modifications. For example, some of our findings have motivated a modification to the fine-tuning protocol, which leads to improved downstream ASR performance in the very low-resource setting. 

Our analyses focus on representations extracted locally (over a frame/phone/word), so it does not measure the infor-mation delocalization that may be happening as a result of the self-attention layers. We leave in-depth analysis of self-attention to future work. Additional future directions include applying the same analytical tools to additional models with different architectures or training objectives, and further studying the implications for additional downstream tasks.

\vspace{.05in}
\noindent\textbf{Acknowledgements.} We thank Shane Settle for providing the AGWEs, and David Yunis, Puyuan Peng, and Shubham Toshniwal for help with preliminary experiments and ideation. This research was funded by NSF award IIS-1816627, by Air Force Office of Scientific Research award FA9550-18-1-0166, and by an AWS Machine Learning Research Award.

\bibliographystyle{IEEEbib}
\bibliography{strings,refs}

\end{document}